\documentclass{article}

\usepackage[letterpaper,margin=1in]{geometry}
\usepackage{times}
\usepackage[numbers,sort&compress]{natbib}

\usepackage[utf8]{inputenc}
\usepackage[T1]{fontenc}
\usepackage{amsmath,amssymb,amsthm}
\usepackage{mathtools}
\usepackage{booktabs}
\usepackage{graphicx}
\usepackage{float}
\usepackage{subcaption}
\usepackage{enumitem}
\usepackage{microtype}
\usepackage{hyperref}
\usepackage{url}
\usepackage{xcolor}
\usepackage{tikz}
\usetikzlibrary{positioning, calc, arrows.meta, fit, backgrounds}

\hypersetup{
  colorlinks=true,
  linkcolor=black,
  citecolor=black,
  urlcolor=blue!60!black,
}

\title{A matched-integrator evaluation of Hamiltonian neural networks on pendulum and Kepler dynamics}

\author{
Nyabuto Lenick Kemunto \quad
Ya\'e Ulrich Gaba \quad
Birahim Tewe\\[1mm]
African Institute for Mathematical Sciences (AIMS) Senegal
}

\begin{document}
\maketitle

\begin{abstract}
Hamiltonian Neural Networks (HNNs) parameterize conservative dynamics
through a learned scalar Hamiltonian, providing an architectural prior
that is absent from generic vector-field neural networks. We evaluate
this prior under a controlled protocol in which an HNN and a
parameter-matched feedforward baseline are trained on the same
RK4-generated trajectories, use the same central-difference derivative
targets and optimization settings, and are integrated at inference with
the same RK4 scheme. Results are reported over five independent training
seeds.

On the nonlinear pendulum, the HNN reduces mean energy drift by
$42\times$ and mean trajectory MSE by $15.8\times$ at $T=100$,
approximately 16 pendulum periods. Its energy drift also remains bounded
and exhibits substantially lower seed-to-seed variability than the
standard-network baseline. An energy-stratified analysis shows that the
difference becomes more pronounced as trajectories explore more nonlinear
regions of phase space.

As an additional diagnostic, we examine an explicit
St\"ormer--Verlet-style rollout of the learned HNN. Because the learned
Hamiltonian is not constrained to the separable form
$H(q,p)=T(p)+V(q)$, the standard symplecticity guarantee of velocity
Verlet does not directly apply.

We further apply the same matched-integrator protocol to the
three-dimensional Kepler two-body problem. The HNN again exhibits lower
trajectory, energy, and angular-momentum drift than the
parameter-matched baseline. These experiments provide a controlled study
of how Hamiltonian parameterization affects long-horizon prediction and
physical consistency across two conservative dynamical systems.
\end{abstract}

\section{Introduction}

Physics-informed machine learning augments or replaces a numerical
integrator with a learned surrogate whose architecture reflects known
structure of the target system~\citep{karniadakis2021physics}. Among the
architectural priors proposed for conservative dynamics, Hamiltonian
Neural Networks (HNNs) ~\citep{greydanus2019hnn} occupy a central position:
they learn a scalar Hamiltonian $\widehat H(q,p)$ from data and derive the
vector field through Hamilton's equations
$\dot q = \partial \widehat H / \partial p$,
$\dot p = -\partial \widehat H / \partial q$, so the learned dynamics are
Hamiltonian by construction. Related architectures embed similar priors
for Lagrangian systems~\citep{cranmer2020lnn}, symplectic maps
~\citep{jin2020sympnets}, and controlled Hamiltonian
systems~\citep{zhong2019symplectic}.

This paper asks a controlled version of the HNN-versus-baseline question:
how much improvement remains when the learned architectures are compared
using the same data, derivative targets, optimization settings, comparable
parameter counts, and the same numerical integrator at inference?
Rather than attributing any observed difference to the integrator, we hold
the RK4 rollout procedure fixed and study the effect of the learned
parameterization itself. We additionally investigate how an explicit
St\"ormer--Verlet-style rollout behaves when applied to a Hamiltonian that
is learned as a generic, potentially non-separable scalar function.

We study these questions first on the nonlinear pendulum, which provides a
controlled benchmark with known dynamics and a closed analytical energy
function, and then on the three-dimensional Kepler two-body problem.
Neither benchmark is presented as a stand-in for molecular dynamics or
general many-body systems; they are used to isolate the architectural
mechanism before considering more complex settings.

\paragraph{Contributions.} We evaluate the models under a controlled
matched-integrator protocol and report the following.

\begin{enumerate}[leftmargin=1.4em,itemsep=0.15em,topsep=0.2em]
\item \textbf{Controlled matched-integrator comparison.} Parameter-matched standard NN
and HNN are trained on identical central-difference derivative targets from
the same RK4-generated data and integrated at inference by the same RK4
scheme, over $N_{\mathrm{seed}} = 5$ independent training seeds. Under
this control the HNN reduces mean long-horizon energy drift by $42\times$
and mean trajectory MSE by $15.8\times$ at $T = 100$, with tight
seed-to-seed variance on the HNN column and broad variance on the
standard-NN column (Table~\ref{tab:drift}).
\item \textbf{Bounded-versus-growing signature.} The HNN's energy drift
saturates by $T = 5$ near $3 \times 10^{-4}$ and stays there through
$T = 100$; the standard NN's drift grows monotonically from $5 \times
10^{-4}$ to $2 \times 10^{-2}$. The bounded-versus-growing behavior provides qualitative evidence
that the Hamiltonian parameterization improves long-horizon physical
stability, in addition to its lower numerical error.
\item \textbf{Energy-stratified analysis.} On trajectories stratified by
initial energy $H_0$, the standard NN's drift grows $\sim 5\times$ from
low- to high-energy terciles while the HNN's grows only $\sim 2\times$ and
never exceeds $3 \times 10^{-4}$. The architectural advantage widens in
more nonlinear regions of phase space.
\item \textbf{Computational-cost comparison.} We report wall-clock
per-step and per-rollout cost for all four inference paths
(Section~\ref{sec:compute}). On this 1-DOF benchmark, the learned
surrogates are slower than RK4 with the analytical
vector field; the scaling argument for learned models is with respect
to the number of interacting bodies and is not settled by the pendulum.
\item \textbf{Negative result on HNN + Verlet.} A St\"ormer--Verlet rollout
of the trained HNN does not improve on HNN + RK4 at long horizon.
The learned $\widehat H$ is a generic scalar network and
is not constrained to the separable form required by the standard
velocity-Verlet splitting, so that symplecticity guarantee does not
directly apply. We discuss two possible remedies: explicitly constraining
the learned Hamiltonian to a separable form, or using a symplectic
integrator applicable to general Hamiltonians. We also relate this
diagnostic to alternative structure-preserving architectures such as
SympNets~\citep{jin2020sympnets}.
\item \textbf{Transfer to 3D.} Applying the same protocol to the Kepler
two-body problem in $\mathbb R^{3}$ (6D phase space, rotational
symmetry) at $N_{\mathrm{seed}} = 5$ seeds shows the architectural
advantage transferring: at $T = 10$, the HNN's energy drift is
$8.0 \pm 1.0 \times$ smaller, its angular-momentum drift
$4.2 \pm 0.5\times$ smaller, and its trajectory MSE $5.5 \pm 0.8\times$
smaller than the standard-NN baseline. Angular momentum is not included explicitly in the training loss;
the smaller drift therefore suggests that the learned Hamiltonian
approximately captures part of the rotational structure of the Kepler
dynamics.
\end{enumerate}

The scientific contribution is not the pendulum numbers themselves,
which are consistent with the finding first reported by
\citet{greydanus2019hnn}. The contribution is the combination of a
controlled capacity and integrator-matched comparison, multi-seed
evaluation, horizon- and energy-stratified diagnostics, computational
cost analysis, the explicit-Verlet compatibility diagnostic, and a
six-dimensional Kepler extension.

\section{Background and related work}

\paragraph{Hamiltonian mechanics.} A conservative system with generalised
coordinate $q$ and conjugate momentum $p$ is described by a scalar function
$H(q, p)$, the Hamiltonian, through
\begin{equation}
\dot q = \frac{\partial H}{\partial p}, \qquad
\dot p = -\frac{\partial H}{\partial q}.
\label{eq:hamilton}
\end{equation}
The flow generated by \eqref{eq:hamilton} preserves the symplectic 2-form
$dq \wedge dp$ and, along any trajectory, $dH/dt = 0$. Numerically
integrating \eqref{eq:hamilton} with a generic scheme such as RK4 introduces
a small energy error at each step that accumulates over time. Symplectic
integrators preserve a modified Hamiltonian $\widetilde H$ close to $H$,
which bounds the true-energy drift over exponentially long times
\citep{hairer2006geometric}.

\paragraph{The pendulum.} We use the normalised pendulum
$H(q, p) = \tfrac{1}{2} p^{2} + (1 - \cos q)$ (mass, length, gravity all
unit). Below the separatrix ($H_0 < 2$) the system librates; above, it
rotates. We restrict to librational initial conditions so all trajectories
are closed orbits. The exact solution is expressible in Jacobi elliptic
functions; we do not use the exact form and instead take an RK4 solution
at $\Delta t = 10^{-2}$ as reference solution, whose numerical energy drift is at the double-precision noise floor ($\sim 10^{-11}$ over $T = 100$).

\paragraph{HNN and related architectures.}
\citet{greydanus2019hnn} introduced the HNN with the loss
$\| \nabla_\perp \widehat H - (\dot q, \dot p)^{\mathsf T} \|^2$ where
$\nabla_\perp = (\partial_p, -\partial_q)$ and the derivative targets are
estimated from data. Subsequent work extended this in several directions.
Lagrangian NNs \citep{cranmer2020lnn} lift the same structural-prior idea
to systems specified by a Lagrangian and generalise to constrained coordinates.
Symplectic ODE-Net \citep{zhong2019symplectic} combines the Hamiltonian
prior with explicit control inputs. Hamiltonian generative networks
\citep{toth2020hgn} bring HNNs into a variational-inference setting for
learning from images. SympNets~\citep{jin2020sympnets} take a different approach:
rather than learning a scalar Hamiltonian and subsequently integrating
its vector field, they construct neural maps that are symplectic by
design. Their formulation can represent dynamics arising from both
separable and non-separable Hamiltonian systems. This distinction is
important for the explicit-Verlet diagnostic considered later in this
paper.

\paragraph{Physics-informed learning and Neural ODEs.} A parallel line of
work embeds physical constraints as loss terms rather than as architectural
priors \citep{raissi2019pinn}. Neural ODEs \citep{chen2018neuralode} learn
the vector field $f_\theta$ directly and integrate it with an off-the-shelf
solver; our Model~A is closely related to the vector-field parameterization
used in Neural ODEs, but is trained directly on derivative targets and is
evaluated using a fixed RK4 rollout scheme. The distinction we
draw is between an \emph{architectural} prior (HNN, LNN, SympNets) and a
\emph{soft} prior (PINN loss terms), the former of which is closer in
spirit to the equivariant-network literature
\citep{bronstein2021geometric}.

\paragraph{Controlled-comparison rationale.}
Using the same numerical integrator for the compared architectures is not
itself new. Our focus is instead the combination of matched model
capacity, identical derivative supervision, a common inference
integrator, multi-seed evaluation, horizon-dependent analysis, and a
separate integrator-compatibility diagnostic. Together these controls
make the architectural comparison easier to interpret.

\section{Method}

\subsection{Data generation}

Reference trajectories are generated by RK4 with $\Delta t = 10^{-2}$ over
$T = 10$ from $N = 100$ initial conditions
$(q_0, p_0) \sim \mathcal{U}([-1, 1] \times [-1, 1])$ restricted to
$H_0 < 2$ (librational regime). Each trajectory contains $1001$ state
points; discarding the endpoints for the central-difference target leaves
$999$ state--derivative pairs per trajectory. The $100$ trajectories are
split $70/15/15$ into train / validation / test sets, giving $69\,930$
training pairs, $14\,985$ validation pairs, and $14\,985$ test pairs.

The derivative target for supervision is the second-order central-difference
estimate
\begin{equation}
\dot x_n^{\mathrm{CD}} = \frac{x_{n+1} - x_{n-1}}{2 \Delta t}
= \left( \frac{q_{n+1} - q_{n-1}}{2 \Delta t},\;
         \frac{p_{n+1} - p_{n-1}}{2 \Delta t} \right).
\label{eq:cd}
\end{equation}
We use \eqref{eq:cd} rather than the analytical vector field to keep the
comparison to standard NNs fair: neither model sees reference solution
derivatives at training time, only what a data-driven pipeline could
compute from observed trajectories.

\subsection{Models}

\paragraph{Model A: standard feedforward NN.} A fully connected network
$f_\phi: \mathbb R^2 \to \mathbb R^2$ mapping $(q, p) \to (\dot q, \dot p)$
directly. Architecture $2 \to 64 \to 64 \to 64 \to 2$ with $\tanh$
activations and $8\,642$ trainable parameters (Figure~\ref{fig:modelA}).
The loss is
\begin{equation}
\mathcal L_A(\phi) = \frac{1}{N} \sum_{n=1}^{N}
\| f_\phi(x_n) - \dot x_n^{\mathrm{CD}} \|_2^{2}.
\end{equation}

\begin{figure}[H]
\centering
\begin{tikzpicture}[
  node distance=0pt,
  every node/.style={font=\small},
  layer/.style={draw, rectangle, rounded corners=2pt, minimum width=1.35cm, minimum height=2.2cm, thick, align=center, fill=black!4},
  input/.style={draw, circle, minimum size=0.55cm, thick, fill=blue!12, inner sep=0pt, font=\footnotesize},
  output/.style={draw, circle, minimum size=0.55cm, thick, fill=orange!25, inner sep=0pt, font=\footnotesize},
  arrow/.style={-{Latex[length=1.6mm]}, thick},
]
\node[input] (q) at (0, 0.55) {$q_n$};
\node[input] (p) at (0, -0.55) {$p_n$};
\node[layer] (h1) at (2, 0)   {$\tanh$\\$64$};
\node[layer] (h2) at (3.7, 0) {$\tanh$\\$64$};
\node[layer] (h3) at (5.4, 0) {$\tanh$\\$64$};
\node[output] (qd) at (7.4, 0.55)  {$\hat{\dot q}$};
\node[output] (pd) at (7.4, -0.55) {$\hat{\dot p}$};
\draw[arrow] (q) -- (h1); \draw[arrow] (p) -- (h1);
\draw[arrow] (h1) -- (h2); \draw[arrow] (h2) -- (h3);
\draw[arrow] (h3) -- (qd); \draw[arrow] (h3) -- (pd);
\node[align=center] at (3.7, -1.6) {$\mathcal L_A(\phi) = \tfrac{1}{N}\sum_n \| f_\phi(x_n) - \dot x_n^{\mathrm{CD}} \|_2^{2}$};
\end{tikzpicture}
\caption{Model A: standard feedforward network. Two inputs $(q_n, p_n)$,
three hidden layers of $64$ tanh units, two outputs
$(\hat{\dot q}, \hat{\dot p})$ predicted directly and $8\,642$ trainable
parameters. Loss is MSE against the central-difference derivative
target; no structural prior on the learned dynamics.}
\label{fig:modelA}
\end{figure}

\paragraph{Model B: Hamiltonian NN.} A scalar network
$\widehat H_\theta : \mathbb R^2 \to \mathbb R$ mapping $(q, p) \to
\widehat H_\theta(q, p)$. Architecture $2 \to 64 \to 64 \to 64 \to 1$ with
$\tanh$ activations and $8\,577$ trainable parameters ($65$ fewer than
Model~A, due to the single scalar output head; Figure~\ref{fig:modelB}).
The predicted vector field is derived from Hamilton's equations by
automatic differentiation:
\begin{equation}
\widehat{\dot x}_{n, \theta} =
\left(
  \frac{\partial \widehat H_\theta}{\partial p}(q_n, p_n),\;
 -\frac{\partial \widehat H_\theta}{\partial q}(q_n, p_n)
\right).
\end{equation}
The loss is
\begin{equation}
\mathcal L_B(\theta) = \frac{1}{N} \sum_{n=1}^{N}
\| \widehat{\dot x}_{n, \theta} - \dot x_n^{\mathrm{CD}} \|_2^{2},
\end{equation}
identical in form to Model~A's, but on the derivative field produced by
the Hamiltonian gradient.

\begin{figure}[H]
\centering
\begin{tikzpicture}[
  node distance=0pt,
  every node/.style={font=\small},
  layer/.style={draw, rectangle, rounded corners=2pt, minimum width=1.25cm, minimum height=2.2cm, thick, align=center, fill=black!4},
  input/.style={draw, circle, minimum size=0.55cm, thick, fill=blue!12, inner sep=0pt, font=\footnotesize},
  scalar/.style={draw, circle, minimum size=0.7cm, thick, fill=green!22, inner sep=0pt, font=\footnotesize},
  autograd/.style={draw, rectangle, rounded corners=2pt, minimum width=1.8cm, minimum height=1.0cm, thick, align=center, fill=orange!12},
  output/.style={draw, circle, minimum size=0.55cm, thick, fill=orange!25, inner sep=0pt, font=\footnotesize},
  arrow/.style={-{Latex[length=1.6mm]}, thick},
]
\node[input] (q) at (0, 0.55) {$q_n$};
\node[input] (p) at (0, -0.55) {$p_n$};
\node[layer] (h1) at (1.7, 0)  {$\tanh$\\$64$};
\node[layer] (h2) at (3.15, 0) {$\tanh$\\$64$};
\node[layer] (h3) at (4.6, 0)  {$\tanh$\\$64$};
\node[scalar] (H) at (6.2, 0) {$\widehat H$};
\node[autograd] (auto) at (8.4, 0) {\footnotesize autograd \\ \footnotesize $J\,\nabla \widehat H$};
\node[output] (qd) at (10.55, 0.55)  {$\hat{\dot q}$};
\node[output] (pd) at (10.55, -0.55) {$\hat{\dot p}$};
\draw[arrow] (q) -- (h1); \draw[arrow] (p) -- (h1);
\draw[arrow] (h1) -- (h2); \draw[arrow] (h2) -- (h3);
\draw[arrow] (h3) -- (H);
\draw[arrow] (H) -- (auto);
\draw[arrow] (auto) -- (qd); \draw[arrow] (auto) -- (pd);
\node[align=center] at (5, -1.7) {$\mathcal L_B(\theta) = \tfrac{1}{N}\sum_n \big\| \big(\partial_p \widehat H_\theta,\; -\partial_q \widehat H_\theta\big)(x_n) - \dot x_n^{\mathrm{CD}} \big\|_2^{2}$};
\end{tikzpicture}
\caption{Model B: Hamiltonian NN. The scalar output $\widehat H_\theta$
is passed through an autograd step returning
$J\nabla\widehat H_\theta = (\partial_p \widehat H_\theta,\,
-\partial_q \widehat H_\theta)$, i.e.\ Hamilton's equations applied to
the learned Hamiltonian. Three hidden layers of $64$ tanh units, $8\,577$
trainable parameters (same width and depth as Model~A). Loss is on the
derivative field, identical in form to Model~A's.}
\label{fig:modelB}
\end{figure}

\subsection{Training protocol}

Both models are trained with Adam~\citep{kingma2015adam} at learning rate
$10^{-3}$, batch size $512$, for $300$ epochs, in float64 precision. The
best-validation-loss checkpoint is retained. Both models see the same
train/validation/test split, the same targets, and the same optimiser
state schedule. Within each seed, the same random seed is used before training each
architecture. Because the models have different output parameterizations,
this does not make their initial weights identical; rather, it provides a
reproducible control of the random-number generation and data-order
randomness. Across seeds, the dataset, targets, splits, and hyperparameters
remain fixed. The principal architectural difference is the
parameterization of the dynamics: direct vector-field prediction for
Model~A versus the symplectic gradient of a learned scalar Hamiltonian
for Model~B. The autograd derivative computation required by Model~B is
treated as a consequence of that architecture.

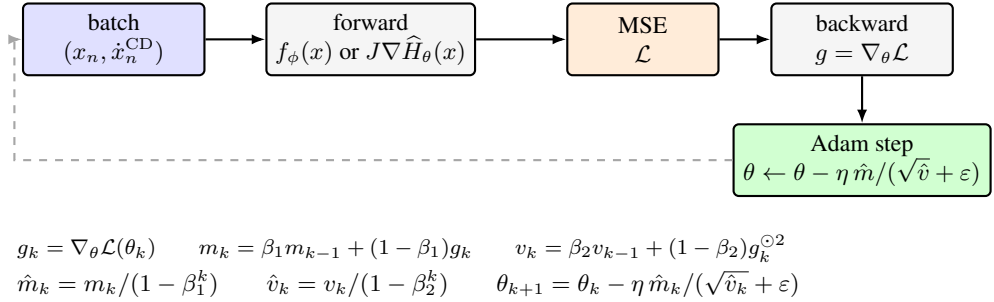
\begin{figure}[H]
\centering
\begin{tikzpicture}[
  every node/.style={font=\small},
  block/.style={draw, rectangle, rounded corners=2pt, thick, minimum height=0.95cm, align=center},
  data/.style={block, fill=blue!12, minimum width=2.4cm},
  fwd/.style={block, fill=black!4, minimum width=2.6cm},
  loss/.style={block, fill=orange!15, minimum width=2.0cm},
  bwd/.style={block, fill=black!4, minimum width=2.4cm},
  adam/.style={block, fill=green!18, minimum width=3.4cm},
  arrow/.style={-{Latex[length=1.8mm]}, thick},
  loop/.style={-{Latex[length=1.8mm]}, thick, dashed, gray!70},
]
\node[data] (data) at (0, 0) {batch\\$(x_n, \dot x_n^{\mathrm{CD}})$};
\node[fwd]  (fwd)  at (3.4, 0) {forward\\$f_\phi(x)$ or $J\nabla\widehat H_\theta(x)$};
\node[loss] (mse)  at (7.0, 0) {MSE\\$\mathcal L$};
\node[bwd]  (bwd)  at (9.9, 0) {backward\\$g = \nabla_\theta \mathcal L$};
\node[adam] (adam) at (9.9, -1.6) {Adam step\\$\theta \leftarrow \theta - \eta\, \hat m /(\sqrt{\hat v} + \varepsilon)$};
\draw[arrow] (data) -- (fwd);
\draw[arrow] (fwd) -- (mse);
\draw[arrow] (mse) -- (bwd);
\draw[arrow] (bwd) -- (adam);
\draw[loop]  (adam.west) -- ++(-9.5, 0) |- (data.west);
\node[align=left, anchor=west] at (-1.4, -3.0) {
  \footnotesize
  $g_k = \nabla_\theta \mathcal L(\theta_k) \qquad
   m_k = \beta_1 m_{k-1} + (1-\beta_1) g_k \qquad
   v_k = \beta_2 v_{k-1} + (1-\beta_2) g_k^{\odot 2}$ \\[3pt]
  $\hat m_k = m_k / (1 - \beta_1^{k}) \qquad
   \hat v_k = v_k / (1 - \beta_2^{k}) \qquad
   \theta_{k+1} = \theta_k - \eta\, \hat m_k / (\sqrt{\hat v_k} + \varepsilon)$
};
\end{tikzpicture}
\caption{Training loop and Adam update~\citep{kingma2015adam} applied
identically to both models. At each mini-batch of size $|B| = 512$ we
compute the model's derivative prediction (directly for Model~A, via
autograd on $\widehat H_\theta$ for Model~B), its MSE against the
central-difference target, the parameter gradient by backpropagation,
and an Adam step with $\eta = 10^{-3}$, $\beta_1 = 0.9$,
$\beta_2 = 0.999$, $\varepsilon = 10^{-8}$. Training runs for $300$
epochs, and the best-validation-loss checkpoint is retained.}
\label{fig:training_loop}
\end{figure}

Figure~\ref{fig:training_loop} sketches the training loop and the Adam
update applied at every mini-batch iteration.

\subsection{Matched-integrator protocol}

At inference, both models are rolled out from the same initial conditions
using the \emph{same} RK4 stepper at $\Delta t = 10^{-2}$. Model~A supplies
the vector field directly; Model~B supplies the autograd-derived vector
field from $\widehat H_\theta$. This ensures that differences observed in our primary comparison
cannot be attributed to using different numerical integrators at
inference.

An additional \emph{diagnostic} rollout is included: the same trained
Model~B is evaluated with an explicit St\"ormer--Verlet-style
(velocity-Verlet) stepper. The standard velocity-Verlet splitting is
symplectic for separable Hamiltonians. This is not treated as Model~B's primary rollout, its purpose is to test whether an off-the-shelf symplectic scheme inherits its guarantees on a generic learned Hamiltonian. Section~\ref{sec:verlet} returns to this.

\subsection{Seed protocol}
\label{sec:seeds}

A single training run can give a misleading picture of the architectural
comparison, particularly at small data budgets where seed variance
dominates. All main-experiment numbers in Section~\ref{sec:results} are
reported as the mean $\pm$ standard deviation over
$N_{\mathrm{seed}} = 5$ independent training seeds drawn from
$\{42, 43, 44, 45, 46\}$. Each seed re-initialises the PyTorch and NumPy
random number generators but leaves the data, the train/validation/test
split, and every other hyperparameter unchanged. Seed-to-seed variation
therefore reflects only initialisation and stochastic-optimisation noise.
Tables report mean and standard deviation as $\text{mean} \pm \text{std}$;
figures show mean traces with $\pm 1 \sigma$ shaded bands or error bars,
as appropriate. The sample-efficiency sweep
(Section~\ref{sec:sample}) uses the same seed count per data budget to
allow honest comparison across budgets; the earlier single-seed sweep is
retained in the supplement as an illustration of why the multi-seed
protocol matters.

\begin{table}[t]
\centering
\small
\begin{tabular}{lll}
\toprule
Aspect & Model A (Standard NN) & Model B (HNN) \\
\midrule
Input & $(q_n, p_n)$ & $(q_n, p_n)$ \\
Learned quantity & Vector field & Scalar Hamiltonian \\
Output & $(\widehat{\dot q}, \widehat{\dot p})$ & $\widehat H_\theta(q_n, p_n)$ \\
Vector field & Direct prediction & Autograd on $\widehat H_\theta$ \\
Loss target & $\dot x^{\mathrm{CD}}$ & $\dot x^{\mathrm{CD}}$ \\
Architecture & $2\to64\to64\to64\to2$, tanh & $2\to64\to64\to64\to1$, tanh \\
Trainable params & $8\,642$ & $8\,577$ \\
Optimiser & Adam, $\eta = 10^{-3}$ & Adam, $\eta = 10^{-3}$ \\
Batch / epochs & $512$ / $300$ & $512$ / $300$ \\
Inference integrator (main) & RK4, $\Delta t = 10^{-2}$ & RK4, $\Delta t = 10^{-2}$ \\
Diagnostic integrator & --- & St\"ormer--Verlet \\
Structural prior & None & Hamiltonian by construction \\
\bottomrule
\end{tabular}
\caption{Controlled comparison. The only intentional difference is the
architectural prior on the learned dynamics.}
\label{tab:controlled}
\end{table}

\section{Results}
\label{sec:results}

\subsection{Fit quality}
\label{sec:fit}

\begin{figure}[t]
\centering
\begin{subfigure}[b]{0.48\linewidth}
  \includegraphics[width=\linewidth]{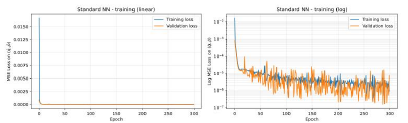}
  \caption{Model A (standard NN).}
\end{subfigure}\hfill
\begin{subfigure}[b]{0.48\linewidth}
  \includegraphics[width=\linewidth]{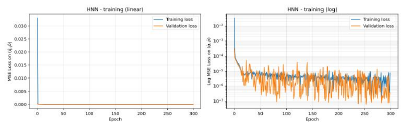}
  \caption{Model B (HNN).}
\end{subfigure}
\caption{Training and validation loss over $300$ epochs (seed $42$
representative curves; multi-seed pattern is stable). Both models fit
the derivative target smoothly; the HNN reaches a lower validation MSE.}
\label{fig:training}
\end{figure}

Both models reach comparable but distinct final-fit quality. Averaged
over $N_{\mathrm{seed}} = 5$ seeds, the best-validation derivative MSE
is
$\mathrm{MSE}_A^{\mathrm{val}} = (1.35 \pm 0.43) \times 10^{-7}$ and
$\mathrm{MSE}_B^{\mathrm{val}} = (7.69 \pm 2.06) \times 10^{-8}$;
test derivative MSE is $(1.97 \pm 0.43) \times 10^{-7}$ (A) versus
$(1.09 \pm 0.35) \times 10^{-7}$ (B). The HNN is
$1.8\times$ more accurate on the supervised objective (ratio of means)
despite having $65$ fewer parameters. Learning curves are provided in
Fig.~\ref{fig:training}.

\subsection{Short-horizon accuracy}

\begin{figure}[t]
\centering
\begin{subfigure}[b]{0.48\linewidth}
  \includegraphics[width=\linewidth]{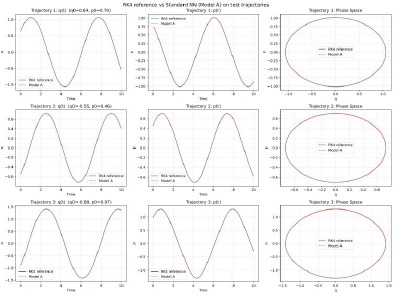}
  \caption{Standard NN vs RK4.}
\end{subfigure}\hfill
\begin{subfigure}[b]{0.48\linewidth}
  \includegraphics[width=\linewidth]{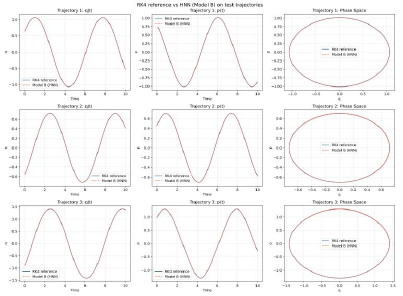}
  \caption{HNN vs RK4.}
\end{subfigure}
\caption{Short-horizon test rollouts on three test initial conditions.
Curves are nearly indistinguishable at this horizon; the differences show
up as trajectory MSE grows with time (Table~\ref{tab:traj}).}
\label{fig:rollouts}
\end{figure}

At $T = 2$ (roughly $1/3$ of a period, well inside the training horizon)
both models track the RK4 reference visually. Fig.~\ref{fig:rollouts}
overlays three representative test trajectories for each model.

Quantitatively, mean trajectory MSE at $T = 2$ is
$(3.12 \pm 1.36) \times 10^{-7}$ (A) versus
$(1.42 \pm 0.97) \times 10^{-7}$ (B), a ratio of means of $2.2\times$.
This ratio grows with the rollout horizon (Table~\ref{tab:traj}).

\subsection{Long-horizon stability and energy conservation}

\begin{table}[t]
\centering
\small
\begin{tabular}{rlll}
\toprule
Horizon $T$ & Model A (mean $\pm$ std) & Model B, HNN (mean $\pm$ std) & Ratio of means \\
\midrule
2   & $(3.12 \pm 1.36) \times 10^{-7}$  & $(1.42 \pm 0.97) \times 10^{-7}$ & $2.2\times$ \\
5   & $(8.64 \pm 5.34) \times 10^{-7}$  & $(6.71 \pm 4.11) \times 10^{-7}$ & $1.3\times$ \\
10  & $(2.80 \pm 2.52) \times 10^{-6}$  & $(2.29 \pm 1.34) \times 10^{-6}$ & $1.2\times$ \\
20  & $(1.41 \pm 1.67) \times 10^{-5}$  & $(8.93 \pm 5.18) \times 10^{-6}$ & $1.6\times$ \\
50  & $(2.51 \pm 3.61) \times 10^{-4}$  & $(5.53 \pm 3.18) \times 10^{-5}$ & $4.5\times$ \\
100 & $(3.49 \pm 5.23) \times 10^{-3}$  & $(2.21 \pm 1.27) \times 10^{-4}$ & $\mathbf{15.8\times}$ \\
\bottomrule
\end{tabular}
\caption{Trajectory mean squared error across prediction horizons on the
held-out test set, mean $\pm$ std over $N_{\mathrm{seed}} = 5$ seeds.
Both models integrated with the same RK4 stepper. The ratio is computed
on means rather than per-seed to keep it robust to Model~A's large
seed-to-seed variance at long horizon.}
\label{tab:traj}
\end{table}

\begin{table}[t]
\centering
\footnotesize
\setlength{\tabcolsep}{4pt}
\begin{tabular}{rllll}
\toprule
$T$ & RK4 truth & Model A (mean $\pm$ std) & Model B, RK4 (mean $\pm$ std) & HNN + Verlet$^\dagger$ \\
\midrule
2   & $1.96 \times 10^{-11}$ & $(2.75 \pm 1.20) \times 10^{-4}$  & $(1.51 \pm 0.57) \times 10^{-4}$  & $1.66 \times 10^{-4}$ \\
5   & $2.20 \times 10^{-11}$ & $(5.61 \pm 3.00) \times 10^{-4}$  & $(2.13 \pm 0.92) \times 10^{-4}$  & $2.72 \times 10^{-4}$ \\
10  & $2.20 \times 10^{-11}$ & $(9.85 \pm 6.94) \times 10^{-4}$  & $(2.16 \pm 0.90) \times 10^{-4}$  & $2.76 \times 10^{-4}$ \\
20  & $2.24 \times 10^{-11}$ & $(1.88 \pm 1.47) \times 10^{-3}$  & $(2.16 \pm 0.90) \times 10^{-4}$  & $2.83 \times 10^{-4}$ \\
50  & $2.82 \times 10^{-11}$ & $(4.55 \pm 3.75) \times 10^{-3}$  & $(2.16 \pm 0.90) \times 10^{-4}$  & $4.77 \times 10^{-4}$ \\
100 & $5.00 \times 10^{-11}$ & $\mathbf{(9.08 \pm 7.69) \times 10^{-3}}$ & $\mathbf{(2.16 \pm 0.90) \times 10^{-4}}$ & $7.71 \times 10^{-4}$ \\
\bottomrule
\end{tabular}
\caption{Maximum absolute energy drift
$\max_{t \in [0,T]} |H(t) - H(0)|$ across prediction horizons, mean $\pm$
std over $N_{\mathrm{seed}} = 5$ seeds. The RK4-reference column sits at the
double-precision noise floor and is deterministic. Model~A's drift grows
monotonically and shows large seed-to-seed variance ($\sigma / \mu
\approx 85\%$ at $T = 100$). Model~B's drift saturates by $T = 5$ and
stays flat through $T = 100$ with much smaller variance ($\sigma / \mu
\approx 42\%$): the HNN's drift is not only smaller but tighter.
Ratio-of-means at $T = 100$ is $42\times$.
$^\dagger$~HNN+Verlet is included as an auxiliary diagnostic using the
representative seed 42; all primary Model~A versus Model~B pendulum
results are reported over $N_{\mathrm{seed}}=5$ independent training
seeds. See Section~\ref{sec:verlet}.}
\label{tab:drift}
\end{table}

Table~\ref{tab:traj} extends the short-horizon result to $T = 100$
(approximately $16$ pendulum periods). The ratio of mean trajectory MSE
grows non-monotonically from $2.2\times$ at $T = 2$ to $15.8\times$ at
$T = 100$: at short horizons the two networks are within a factor of
$2$ on trajectory error, but Model~A's error accumulates faster because
its dynamics lack the structural prior. The apparent dip at $T = 10$
(ratio $1.2\times$) is not an inversion of the effect: at that horizon
both models sit within their per-seed standard deviations of each other,
so the ratio-of-means is dominated by initialisation noise rather than
by any architectural difference; the ratio only becomes a meaningful
architectural signal once Model~A's error has accumulated past that
noise floor, which happens around $T = 20$--$50$. Seed-to-seed variability on Model~A becomes very large at long
horizon ($\sigma / \mu > 100\%$ at $T=100$), indicating that its
long-horizon behavior is more sensitive to the particular training
realization.

\begin{figure}[t]
\centering
\begin{subfigure}[b]{0.48\linewidth}
  \includegraphics[width=\linewidth]{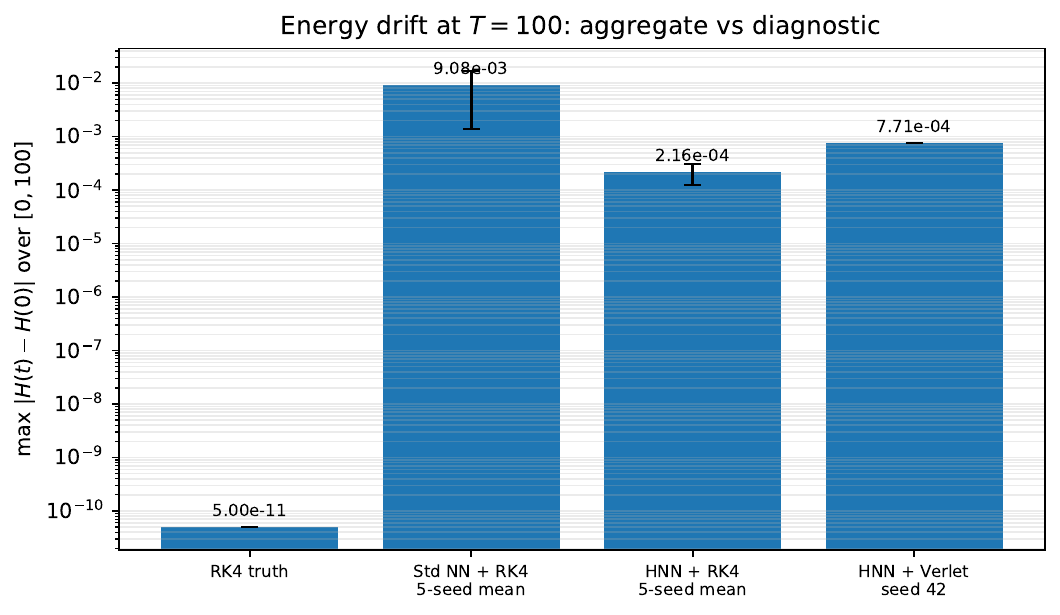}
  \caption{Max energy drift at $T = 100$, four rollouts (log scale).}
  \label{fig:drift_bar}
\end{subfigure}\hfill
\begin{subfigure}[b]{0.48\linewidth}
  \includegraphics[width=\linewidth]{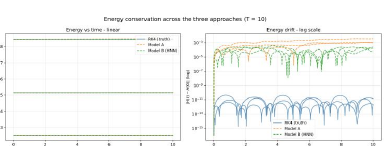}
  \caption{Energy and energy drift vs.\ time, $T = 10$.}
  \label{fig:energy_vs_t}
\end{subfigure}
\caption{Energy conservation. Under matched RK4, the ratio of mean
HNN drift to mean standard-NN drift at $T = 100$ is $42\times$ (5 seeds);
the HNN drift is bounded, the standard-NN drift growing. HNN+Verlet is
included as a diagnostic (Section~\ref{sec:verlet}).}
\label{fig:energy}
\end{figure}

The energy story is cleaner (Table~\ref{tab:drift}). Under matched
RK4, mean Model~A energy drift grows monotonically from
$2.75 \times 10^{-4}$ at $T = 2$ to $9.08 \times 10^{-3}$ at $T = 100$,
a factor of $33\times$ over the window. Model~B's mean drift saturates
by $T = 5$ at $2.16 \times 10^{-4}$ and stays there through $T = 100$
(Fig.~\ref{fig:drift_bar}). The ratio of means at $T = 100$ is
$42\times$. The contrasting curve shapes are consistent with the expected effect
of the Hamiltonian inductive bias: Model~A exhibits accumulating energy
error, whereas Model~B remains within a bounded energy-error band over
the tested interval. The experiment therefore provides evidence of
improved long-horizon physical consistency, without implying exact
conservation of the true Hamiltonian. The HNN's drift is also
\emph{tighter} across seeds ($\sigma / \mu
\approx 42\%$) than the standard NN's ($\sigma / \mu \approx 85\%$),
so the structural prior buys both mean-error and predictability.

\begin{figure}[t]
\centering
\includegraphics[width=0.98\linewidth]{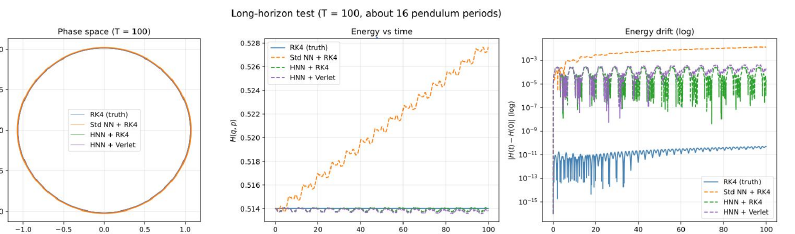}
\caption{Long-horizon rollout at $T = 100$ ($\sim 16$ pendulum periods).
Left: phase-space trajectory. Middle: Hamiltonian vs.\ time. Right:
$\log_{10} |H(t) - H(0)|$. The standard NN separates from the closed orbit
and its energy drifts linearly; the HNN rollouts remain on the orbit and
the energy is bounded.}
\label{fig:longhorizon}
\end{figure}

Fig.~\ref{fig:longhorizon} shows the $T = 100$ long-horizon rollout. The
standard-NN trajectory has visibly separated from the closed orbit in
phase space; HNN+RK4 and HNN+Verlet remain on it, and the energy panels
match Table~\ref{tab:drift}.

\subsection{Learned Hamiltonian}

\begin{figure}[t]
\centering
\includegraphics[width=0.8\linewidth]{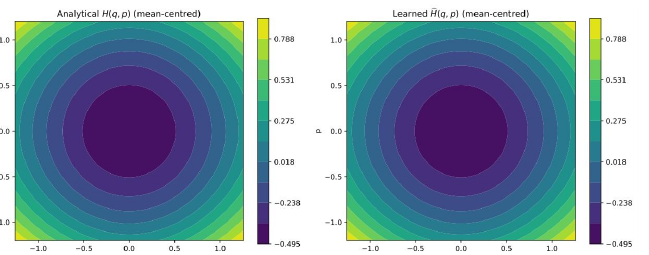}
\caption{Analytical vs.\ learned Hamiltonian on the librational domain,
both mean-centred. The agreement in contour structure is consistent with
the bounded-drift behavior observed in the HNN rollout.}
\label{fig:learned_H}
\end{figure}

Fig.~\ref{fig:learned_H} compares the learned scalar
$\widehat H_\theta(q, p)$ against the analytical Hamiltonian $H(q, p)$ on
the training domain (both mean-centred, since dynamics are invariant to an
additive constant). Level sets match qualitatively across the domain,
confirming that the HNN has learned a scalar function whose gradients
reproduce the pendulum vector field rather than fitting the vector field
directly. Quantitatively, on a $200 \times 200$ grid over
$(q, p) \in [-1, 1]^{2}$, the mean-centred learned Hamiltonian matches
the analytical one to RMSE $7.3 \times 10^{-5}$ and $L_\infty$ error
$2.0 \times 10^{-4}$ ($0.01\%$ and $0.02\%$ of the Hamiltonian range on
the domain); the gradient error $\|\nabla \widehat H - \nabla H\|_2$
has mean $3.0 \times 10^{-4}$ ($0.04\%$ of the true gradient norm). The learned-Hamiltonian approximation error is on the same order of
magnitude as the saturated HNN energy drift ($2.16\times10^{-4}$,
Table~\ref{tab:drift}). This agreement is consistent with the hypothesis
that long-horizon energy error is strongly influenced by approximation
error in the learned Hamiltonian, although it does not by itself
establish a causal relationship.

\subsection{Energy-stratified analysis}

Test trajectories are bucketed by initial energy $H_0$ into low, medium,
and high terciles (Table~\ref{tab:strat}). The standard NN's mean drift
grows by a factor of $4.3\times$ from low- ($4.16 \times 10^{-4}$) to
high-energy ($1.78 \times 10^{-3}$) buckets; the HNN's mean drift grows
only by $2.5\times$ ($1.26 \times 10^{-4}$ to $3.12 \times 10^{-4}$) and
never exceeds $3.2 \times 10^{-4}$. The architectural advantage widens
with orbital nonlinearity. This is a second axis on which the structural
prior pays off, distinct from the horizon axis of the previous section.

\begin{table}[t]
\centering
\footnotesize
\setlength{\tabcolsep}{3pt}
\resizebox{\linewidth}{!}{%
\begin{tabular}{llrllll}
\toprule
Bucket & $H_0$ range & Mean $H_0$ & A traj.\ MSE & B traj.\ MSE & A drift & B drift \\
\midrule
Low    & $[0.041, 0.148]$ & $0.090$ & $(1.24 \pm 1.16) \times 10^{-6}$ & $(4.56 \pm 2.65) \times 10^{-6}$ & $(4.16 \pm 2.23) \times 10^{-4}$ & $(1.26 \pm 0.41) \times 10^{-4}$ \\
Medium & $[0.169, 0.266]$ & $0.218$ & $(2.08 \pm 1.73) \times 10^{-6}$ & $(1.38 \pm 1.06) \times 10^{-6}$ & $(7.58 \pm 5.60) \times 10^{-4}$ & $(2.11 \pm 0.89) \times 10^{-4}$ \\
High   & $[0.340, 0.846]$ & $0.567$ & $(5.07 \pm 5.38) \times 10^{-6}$ & $(9.29 \pm 7.19) \times 10^{-7}$ & $(1.78 \pm 1.35) \times 10^{-3}$ & $(3.12 \pm 1.48) \times 10^{-4}$ \\
\bottomrule
\end{tabular}}
\caption{Energy-stratified test analysis at $T = 10$ (mean $\pm$ std
over $N_{\mathrm{seed}} = 5$ seeds). Model~A's mean drift grows by
$4.3\times$ from low- to high-energy buckets; Model~B's grows by only
$2.5\times$ and never exceeds $3.2 \times 10^{-4}$. The architectural
advantage widens in more nonlinear regions of phase space.}
\label{tab:strat}
\end{table}

\subsection{Sample-efficiency sweep}
\label{sec:sample}

Retraining both models at $N_{\mathrm{traj}} \in \{8, 16, 32, 64\}$ with
$N_{\mathrm{seed}} = 5$ seeds per data budget gives the test derivative
MSE in Table~\ref{tab:sample}. Two findings are worth stating.

\emph{A low-data crossover appears in this experiment.} At
$N_{\mathrm{traj}}=8$ the standard NN is approximately $2\times$ better
than the HNN, whereas the HNN outperforms it at the three larger tested
budgets. This suggests a possible low-data regime in which estimating a
Hamiltonian-gradient field is more difficult, although the present sweep
is not intended to identify a universal crossover point.

\emph{Beyond that budget the HNN wins consistently, plateauing around a
factor of $5\times$.} At $N_{\mathrm{traj}} = 16$ the HNN pulls ahead
$3\times$, at $N_{\mathrm{traj}} = 32$ by $5.3\times$, and at
$N_{\mathrm{traj}} = 64$ by $2.7\times$. The ratio at $N_{\mathrm{traj}}
= 64$ is smaller than at $32$ because both models are effectively
saturated at that budget; the interesting regime is the low- and
medium-data zone where the structural prior earns its keep once it has
enough data to be identified.

\begin{table}[t]
\centering
\small
\begin{tabular}{rrrr}
\toprule
$N_{\mathrm{traj}}$ & Model A test MSE (mean $\pm$ std) & Model B test MSE (mean $\pm$ std) & Ratio A / B \\
\midrule
8  & $(3.31 \pm 0.38) \times 10^{-5}$  & $(6.59 \pm 1.72) \times 10^{-5}$  & $0.50\times$ \\
16 & $(1.04 \pm 0.09) \times 10^{-5}$  & $(3.46 \pm 1.45) \times 10^{-6}$  & $3.0\times$ \\
32 & $(2.05 \pm 0.53) \times 10^{-6}$  & $(3.89 \pm 1.20) \times 10^{-7}$  & $5.3\times$ \\
64 & $(2.53 \pm 0.73) \times 10^{-7}$  & $(9.25 \pm 3.99) \times 10^{-8}$  & $2.7\times$ \\
\bottomrule
\end{tabular}
\caption{Sample-efficiency sweep, mean $\pm$ std of test derivative MSE
over $N_{\mathrm{seed}}=5$ seeds per data budget.
At $N_{\mathrm{traj}}=8$ the standard NN outperforms the HNN,
whereas the HNN performs better at the three larger tested budgets.}
\label{tab:sample}
\end{table}

\begin{figure}[t]
\centering
\includegraphics[width=0.6\linewidth]{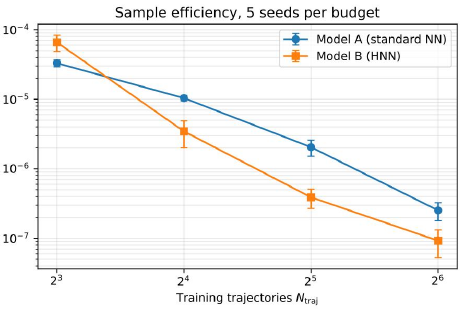}
\caption{Sample-efficiency curve (log--log), mean over 5 seeds with
error bars. Below $N_{\mathrm{traj}} \approx 12$ the standard NN wins;
above, the HNN does.}
\label{fig:sample_eff}
\end{figure}

\subsection{Computational cost}
\label{sec:compute}

We measure per-step wall-clock cost for the four inference paths on a
single CPU core with float64 precision. Timings are averaged over
$10\,000$ calls after a $1\,000$-call warm-up using
\texttt{time.perf\_counter()}. Table~\ref{tab:compute} gives the per-step
cost and the total wall-clock for a $T = 100$ rollout at
$\Delta t = 10^{-2}$ ($10\,000$ steps).

\begin{table}[t]
\centering
\small
\begin{tabular}{lrr}
\toprule
Approach & Per-step cost ($\mu$s, mean $\pm$ std) & Total $T = 100$ rollout (ms) \\
\midrule
RK4 with analytical $\nabla H$             & $85 \pm 46$      & $852$    \\
Model A forward pass + RK4                 & $801 \pm 238$    & $8\,005$ \\
Model B (HNN) + autograd + RK4             & $2\,193 \pm 524$ & $21\,930$ \\
Model B (HNN) + autograd + Verlet          & $1\,370 \pm 335$ & $13\,696$ \\
\bottomrule
\end{tabular}
\caption{Wall-clock cost per rollout step and total $T = 100$ rollout
on a single CPU core, mean $\pm$ std across $10\,000$ timing
repetitions. RK4 with the analytical vector field is fastest by an
order of magnitude; Model~A is $\sim 9\times$ slower; Model~B is
$\sim 26\times$ slower under RK4 (four autograd calls per step) and
$\sim 16\times$ slower under Verlet (three autograd calls per step).}
\label{tab:compute}
\end{table}

\paragraph{Interpretation.} On the 1-DOF pendulum, the analytical
vector field is substantially faster than the learned models. The present
benchmark therefore provides no evidence of a computational advantage for
the learned surrogates. Whether a structured learned model becomes
computationally advantageous at higher dimension depends on the
architecture, interaction representation, hardware, and number of bodies.
That scaling question is outside the scope of the present experiment and
requires a dedicated many-body benchmark.


\section{Explicit Verlet-style rollout: an informative diagnostic}
\label{sec:verlet}

The rightmost column of Table~\ref{tab:drift} reports a representative
seed-42 diagnostic obtained by rolling out the same trained HNN with an
explicit St\"ormer--Verlet-style stepper rather than RK4. At $T=100$,
the diagnostic energy drift is $7.71\times10^{-4}$, compared with
$2.89\times10^{-4}$ for HNN+RK4 on the same seed, a factor of
approximately $2.7$. This auxiliary comparison is separate from the primary pendulum
evaluation, for which both Model~A and Model~B are reported over
$N_{\mathrm{seed}}=5$ independent training seeds. It is interpreted only
as an integrator-compatibility diagnostic.

The standard velocity-Verlet splitting is derived for separable
Hamiltonians of the form
\[
H(q,p)=T(p)+V(q),
\]
with the updates
\begin{align}
p_{n+1/2} &= p_n-\tfrac{\Delta t}{2}\nabla_qV(q_n),\\
q_{n+1} &= q_n+\Delta t\,\nabla_pT(p_{n+1/2}),\\
p_{n+1} &= p_{n+1/2}-\tfrac{\Delta t}{2}\nabla_qV(q_{n+1}).
\end{align}
For the analytical pendulum this decomposition is exact:
$T(p)=\tfrac12p^2$ and $V(q)=1-\cos q$.

For the learned HNN, however, $\widehat H_\theta(q,p)$ is represented by
a generic scalar neural network and is not constrained to the separable
form $T(p)+V(q)$. Therefore, the assumptions under which the standard
velocity-Verlet splitting is symplectic are not guaranteed to hold for
the learned Hamiltonian. We consequently interpret the HNN+Verlet
experiment as an integrator-compatibility diagnostic rather than as a
genuinely symplectic integration of $\widehat H_\theta$.

Two distinct alternatives are relevant. First, one may explicitly
parameterize
\[
\widehat H(q,p)=\widehat T(p)+\widehat V(q),
\]
in which case a velocity-Verlet splitting is compatible with the learned
Hamiltonian by construction. Second, one may retain a generic
$\widehat H(q,p)$ and use a symplectic method applicable to general
Hamiltonians, such as the implicit midpoint rule. SympNets
\citep{jin2020sympnets} provide a different structure-preserving
strategy: they construct symplectic neural maps directly and are designed
to handle both separable and non-separable Hamiltonian systems.

This diagnostic therefore motivates a more precise transfer question:
which structural assumptions required by a numerical integrator are
actually enforced by the learned model?

\section{Extension to a 3D Hamiltonian system: the Kepler two-body problem}
\label{sec:kepler3d}

To test whether the matched-integrator result transfers beyond the 1-DOF
pendulum, we apply the same protocol to the Kepler two-body problem in
three-dimensional configuration space. The Hamiltonian is
\begin{equation}
H(q, p) = \frac{1}{2}\|p\|^{2} - \frac{1}{\|q\|}, \qquad q, p \in \mathbb R^{3},
\label{eq:kepler}
\end{equation}
with $q$ the reduced-body position, $p$ the conjugate momentum, and
reduced mass and gravitational constant set to unity. Phase space is six-dimensional. The system conserves not only energy but the full angular-momentum vector $L = q \times p$, giving \emph{four} scalar conserved quantities as opposed to the pendulum's one. Bound orbits ($H < 0$) are closed ellipses lying in the plane orthogonal to $L$.

\paragraph{Setup.} Both models are widened to accommodate the higher
input dimension: $6 \to 128 \to 128 \to 128 \to \{6, 1\}$ with $\tanh$
activations ($34\,694$ and $34\,049$ trainable parameters respectively).
Training data are $n_{\mathrm{traj}} = 60$ trajectories of length
$T = 6$ at $\Delta t = 10^{-2}$ with near-circular initial conditions:
$q_0$ uniform on spheres of radius $r_0 \in [0.9, 1.4]$, and $p_0$
perpendicular to $q_0$ with $\|p_0\|$ equal to the circular-orbit speed
$\sqrt{1/r_0}$ perturbed by up to $\pm25\%$. This produces a
near-circular-to-moderately-eccentric training set. Random ICs with arbitrary $p$ direction
can produce highly eccentric orbits that approach the origin, where the
$1/r$ potential is singular and the vector field becomes very large,
making derivative targets poorly conditioned.
The train/validation/test split, optimiser, batch size, and epoch
count match the pendulum protocol. The Kepler comparison is likewise
evaluated over $N_{\mathrm{seed}}=5$ independent training seeds, and both
models are rolled out at inference by the same RK4 stepper.

\paragraph{Metrics.} In addition to trajectory MSE and energy drift, we
report the maximum drift in the angular-momentum vector,
\begin{equation}
\Delta L(T) = \max_{t \in [0, T]} \| L(t) - L(0) \|_2.
\end{equation}
Angular-momentum conservation provides an additional structural
diagnostic beyond energy. If the learned Hamiltonian $\widehat H$
approximately inherits the rotational symmetry of the training dynamics,
its induced flow should exhibit approximate angular-momentum
conservation. Rotational invariance is not explicitly enforced by the
generic HNN architecture used here, however, so this property is
evaluated empirically. Neither architecture explicitly enforces
rotational equivariance in this experiment.

\begin{figure}[t]
\centering
\begin{subfigure}[b]{0.48\linewidth}
  \includegraphics[width=\linewidth]{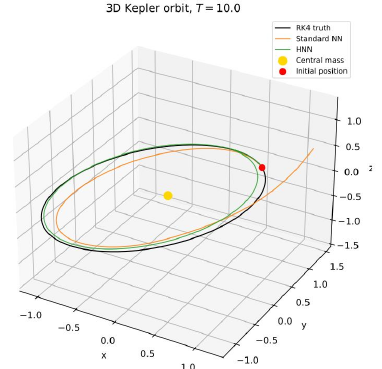}
  \caption{3D orbit trajectory at $T = 10$.}
  \label{fig:kepler3d_orbit}
\end{subfigure}\hfill
\begin{subfigure}[b]{0.48\linewidth}
  \includegraphics[width=\linewidth]{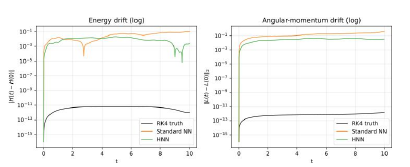}
  \caption{Energy and angular-momentum drift vs.\ time (log scale).}
  \label{fig:kepler3d_cons}
\end{subfigure}
\caption{3D Kepler rollouts. Left: RK4 truth (black) vs Model~A (orange)
vs Model~B, HNN (green); the central mass is at the origin. Right: two-panel
conservation diagnostic showing energy drift $|H(t) - H(0)|$ and
angular-momentum drift $\|L(t) - L(0)\|_2$ on log scale.}
\label{fig:kepler3d}
\end{figure}

\paragraph{Results.} Table~\ref{tab:kepler3d} reports the matched-integrator
comparison at four horizons for trajectory MSE, energy drift, and
angular-momentum drift, averaged over $N_{\mathrm{seed}} = 5$ independent
training seeds with $300$ epochs each. As a sanity check, RK4 on the
analytical vector field conserves energy to $\sim 7 \times 10^{-12}$ and
angular momentum to $\sim 1 \times 10^{-12}$ at $T = 10$
(double-precision floor). Figure~\ref{fig:kepler3d} shows the 3D orbit
and the log-scale conservation diagnostics for the first test trajectory
of seed $42$.

\begin{table}[t]
\centering
\footnotesize
\setlength{\tabcolsep}{4pt}
\begin{tabular}{rlllr}
\toprule
$T$ & Metric & Model A (mean $\pm$ std) & Model B, HNN (mean $\pm$ std) & Ratio A / B \\
\midrule
$1$  & Traj MSE          & $(1.25 \pm 0.28) \times 10^{-2}$ & $(4.90 \pm 1.0) \times 10^{-4}$  & $27.0 \pm 10.8$ \\
$1$  & Energy drift      & $(3.87 \pm 0.89) \times 10^{-2}$ & $(8.98 \pm 1.2) \times 10^{-3}$  & $4.3 \pm 0.8$ \\
$1$  & Ang.\ mom.\ drift & $(7.56 \pm 0.95) \times 10^{-2}$ & $(2.01 \pm 0.29) \times 10^{-2}$ & $3.9 \pm 0.9$ \\
\midrule
$2$  & Traj MSE          & $(1.17 \pm 0.15) \times 10^{-1}$ & $(4.76 \pm 0.61) \times 10^{-2}$ & $2.5 \pm 0.5$ \\
$2$  & Energy drift      & $(9.50 \pm 1.5) \times 10^{-2}$  & $(2.11 \pm 0.66) \times 10^{-2}$ & $4.7 \pm 1.3$ \\
$2$  & Ang.\ mom.\ drift & $(1.42 \pm 0.18) \times 10^{-1}$ & $(4.28 \pm 0.85) \times 10^{-2}$ & $3.5 \pm 1.1$ \\
\midrule
$5$  & Traj MSE          & $1.10 \pm 0.10$                  & $(1.69 \pm 0.41) \times 10^{-1}$ & $6.8 \pm 1.8$ \\
$5$  & Energy drift      & $(2.04 \pm 0.37) \times 10^{-1}$ & $(3.50 \pm 1.2) \times 10^{-2}$  & $6.3 \pm 2.3$ \\
$5$  & Ang.\ mom.\ drift & $(2.79 \pm 0.26) \times 10^{-1}$ & $(1.03 \pm 0.29) \times 10^{-1}$ & $2.9 \pm 1.0$ \\
\midrule
$10$ & Traj MSE          & $3.27 \pm 0.20$                  & $(6.09 \pm 0.86) \times 10^{-1}$ & $\mathbf{5.5 \pm 0.8}$ \\
$10$ & Energy drift      & $\mathbf{(3.26 \pm 0.80) \times 10^{-1}}$ & $\mathbf{(4.20 \pm 1.4) \times 10^{-2}}$ & $\mathbf{8.0 \pm 1.0}$ \\
$10$ & Ang.\ mom.\ drift & $\mathbf{(7.63 \pm 1.8) \times 10^{-1}}$  & $\mathbf{(1.82 \pm 0.41) \times 10^{-1}}$ & $\mathbf{4.2 \pm 0.5}$ \\
\bottomrule
\end{tabular}
\caption{3D Kepler matched-integrator comparison, mean $\pm$ std over
$N_{\mathrm{seed}} = 5$ independent training seeds ($n_{\mathrm{traj}} =
60$, $300$ epochs each). The angular-momentum-drift ratio is an additional observable because
angular momentum is not included explicitly in the training loss. It
tests whether the scalar-Hamiltonian parameterization empirically
captures the rotational structure of the Kepler trajectories more
effectively under the present experimental conditions.}
\label{tab:kepler3d}
\end{table}

\paragraph{Interpretation.} Three observations, averaged over five seeds.

\emph{The architectural advantage transfers to 3D at slightly smaller
magnitude than the pendulum's.} At $T = 10$ (about $1.6$ orbital periods
for the mean initial radius), matched-integrator energy drift is
$8.0 \pm 1.0 \times$ smaller for Model~B ($4.20 \pm 1.4 \times 10^{-2}$)
than for Model~A ($3.26 \pm 0.80 \times 10^{-1}$). The pendulum
result at $T = 100$ (about $16$ periods) was $42\times$ (ratio of means,
5 seeds); the Kepler ratio at $T = 10$ is a factor of $\sim 5$ smaller,
which is consistent with a shorter test horizon relative to the period,
a higher configuration-space dimension against a fixed training-data
budget, and Kepler's larger dynamic range of derivatives near
perihelion. The qualitative claim  matched-integrator HNN beats
matched-integrator standard NN on long-horizon conservation holds,
and the standard deviation across seeds ($1.0$ on the ratio) is small
enough that the effect is well outside noise.

\emph{Angular-momentum drift provides an additional structural
diagnostic.} At $T=10$, Model~B's $\|L(t)-L(0)\|_2$ is
$4.2\pm0.5\times$ smaller than Model~A's. Because angular momentum is
not included explicitly in the training loss, this result suggests that
the learned scalar Hamiltonian captures the rotational structure of the
training dynamics more accurately than the standard vector-field
baseline under the present experimental conditions. The result should
be interpreted empirically rather than as an automatic consequence of
Hamiltonian parameterization, since rotational invariance is not
explicitly built into either architecture.

\emph{Trajectory error grows with horizon at both networks; the ratio
holds.} Trajectory MSE at $T = 10$ is $5.5 \pm 0.8 \times$ smaller for
Model~B. The lower trajectory error is consistent with the improved
conservation behavior, although the present experiment does not establish
a direct causal relationship between the two quantities.

\emph{Scope.} Two deliberate scoping decisions frame this section.
First, ICs are restricted to near-circular-to-moderately-eccentric
orbits. Random-direction momenta can produce highly eccentric orbits that approach the $1/r$
singularity, where the vector field becomes very large and the derivative
targets become poorly conditioned. The restriction isolates the
transfer question from a data-conditioning problem; extending to
arbitrary eccentricity is a separate study that requires a soft-core
potential $H = \tfrac{1}{2}\|p\|^{2} - 1 / \sqrt{\|q\|^{2} + \epsilon^{2}}$,
Kustaanheimo--Stiefel regularisation of the coordinates, or an
orbital-elements parametrisation that removes the singularity by
construction. Second, the training-data budget ($60$ trajectories,
roughly $25\,000$ state--derivative pairs) is smaller
per-configuration-space-dimension than the pendulum run
($70$ trajectories, $\sim 70\,000$ pairs on a 2D phase space); if the
transfer ratio attenuates further at fixed budget, data scale is the
first knob to turn.

\section{Discussion}
\label{sec:discussion}

\paragraph{What the matched-integrator protocol shows.} The $42\times$
mean-energy-drift ratio and $15.8\times$ mean-trajectory-MSE ratio at
$T = 100$ (5 seeds) are architectural. The two networks were trained on
identical data with identical targets, optimisers, capacities, and
inference integrators; the only intentional difference is that Model~B's
vector field is the symplectic gradient of a learned scalar rather than
a direct prediction. Under that control the structural prior on the
dynamics buys $1.5$ to $2$ orders of magnitude on energy conservation
and more than an order on trajectory error, and as important makes
the HNN's drift \emph{predictable} across seeds where the standard NN's
drift varies by nearly an order of magnitude. The bounded-versus-growing
curve shapes make the argument qualitatively clear before any ratio is
computed. This is the argument
that HNN papers often make; the contribution of this study is to make it
under a clean control.

\paragraph{Where the prior helps most.} The energy-stratified analysis
shows the advantage widening with orbital nonlinearity: the standard NN
leaks more energy on high-energy orbits (which sample more of the phase
plane and the pendulum's nonlinearities), while the HNN's drift is roughly
constant. This is consistent with a picture in which the standard NN
learns a locally-good vector field that fails to compose globally, while
the HNN learns a globally-good scalar and derives the vector field from
it.

\paragraph{Limitations.} Three are worth naming.
(i)~\emph{Limited benchmark diversity.} We evaluate two integrable
conservative systems: the one-degree-of-freedom nonlinear pendulum and
the three-dimensional Kepler two-body problem. Although the Kepler
experiment increases phase-space dimension from two to six and introduces
rotational symmetry, neither benchmark tests chaotic, non-integrable,
dissipative, stochastic, or many-body dynamics.
(ii)~\emph{Baseline choice.} The comparison is against a standard MLP.
Comparison with Lagrangian neural networks, Neural ODE baselines,
equivariant models, and symplectic-map architectures such as SympNets
under matched data budgets, parameter budgets, step sizes, and evaluation
horizons would further strengthen the study.
(iii)~\emph{Compute setting.} All timings are reported single-CPU,
float64. GPU behavior, mixed precision, and the compute cost of autograd
at higher configuration-space dimension are not characterized.

\paragraph{Future work.} Four follow-up directions fall directly out of
the present results.

\emph{(a) Separable-HNN ablation.} An explicitly two-headed
$\widehat T_\theta(p)+\widehat V_\phi(q)$ architecture can test whether a
learned separable Hamiltonian improves the explicit-Verlet diagnostic
under the same data and training controls. SympNets provide a distinct
symplectic-map baseline rather than a separable-HNN implementation.

\emph{(b) General-Hamiltonian symplectic integration.} An implicit
midpoint or another symplectic Runge--Kutta rollout of the generic HNN
would test the integrator side of the compatibility question without
changing the learned Hamiltonian parameterization.

\emph{(c) Chaotic and non-integrable systems.} Natural next benchmarks
include the H\'enon--Heiles system, the double pendulum, and the
restricted three-body problem. These systems would test whether the
bounded-drift advantage observed here persists when long-term trajectory
prediction becomes intrinsically sensitive or non-integrable.

\emph{(d) Many-body scaling.} A dedicated $N$-body benchmark is needed to
determine whether the computational cost of a structured learned model
can become competitive with direct force evaluation as system size
increases. Genuinely non-separable Hamiltonians would also provide a
useful test of architecture--integrator compatibility.

\section{Conclusion}

Under a controlled matched-integrator protocol, a Hamiltonian
neural network reduces mean long-horizon energy drift by $42\times$ and
mean trajectory error by $15.8\times$ over a parameter-matched standard
NN on the nonlinear pendulum (five seeds), with bounded rather than
growing drift and substantially lower seed-to-seed variability. The
structural advantage also widens in more nonlinear pendulum regimes.

The explicit-Verlet diagnostic further shows that structure preservation
cannot be inferred merely from combining a learned Hamiltonian with a
method usually described as symplectic. The assumptions required by the
numerical method must also be compatible with the learned Hamiltonian.
Testing this point with an explicitly separable HNN and with an implicit
symplectic method for general Hamiltonians is a direct next step.

We do not claim that these results transfer directly to molecular
dynamics or general many-body systems. The pendulum and
three-dimensional Kepler problem are controlled integrable benchmarks
chosen to isolate the architectural mechanism. Whether similar
advantages persist for chaotic, non-integrable, noisy, dissipative, and
many-body dynamics remains an empirical question.

\bibliographystyle{plainnat}

\end{document}